\documentclass[runningheads]{llncs}
\usepackage{graphicx}
\usepackage{subcaption}
\usepackage{xcolor}
\usepackage{esvect}
\usepackage{booktabs}
\usepackage{amsfonts}
\usepackage{amsmath}
\usepackage{url}
\usepackage[ruled,vlined,linesnumbered]{algorithm2e}

\DeclareGraphicsExtensions{.pdf,.png}

\begin{document}

\title{Contextual Importance and Utility: a Theoretical Foundation \thanks{The work is partially supported by the Wallenberg AI, Autonomous Systems and Software Program (WASP) funded by the Knut and Alice Wallenberg Foundation.
}}
%
%
 \author{Kary Fr\"amling\inst{1,2}\orcidID{0000-0002-8078-5172}
}
\authorrunning{K. Fr\"amling}
%
\institute{Department of Computing Science, Ume\r{a} University \\ Mit-huset, 901 87 Ume\r{a}, Sweden \\
 \email{kary.framling@umu.se} \and
 Department of Computer Science, Aalto University \\
 Konemiehentie 1, 02150 Espoo, Finland 
}
\maketitle              
\begin{abstract}

This paper provides new theory to support to the eXplainable AI (XAI) method Contextual Importance and Utility (CIU). CIU arithmetic is based on the concepts of Multi-Attribute Utility Theory, which gives CIU a solid theoretical foundation. The novel concept of \textit{contextual influence} is also defined, which makes it possible to compare CIU directly with so-called additive feature attribution (AFA) methods for model-agnostic outcome explanation. One key takeaway is that the `influence' concept used by AFA methods is inadequate for outcome explanation purposes even for simple models to explain. Experiments with simple models show that explanations using contextual importance (CI) and contextual utility (CU) produce explanations where influence-based methods fail. It is also shown that CI and CU guarantees explanation faithfulness towards the explained model. 

\begin{sloppypar}
\end{sloppypar}

\end{abstract}
\section{Introduction}



\begin{sloppypar}
Contextual Importance and Utility (CIU) was originally proposed by Kary Fr\"{a}mling in 1995 in a context of Multiple Criteria Decision Making (MCDM). MCDM is a domain where mathematical models are used as Decision Support Systems (DSS) for human decision makers. No matter what model is being used for the DSS, it is crucial that the recommendations or outcome of the DSS can be presented in ways that are understandable for the decision makers, as well as for the people who might be affected by the decisions. 
CIU is model-agnostic and provides uniform explanation concepts for all possible DSS models, ranging from linear models such as the weighted sum, to rule-based systems, decision trees, fuzzy systems, neural networks and any machine learning-based models.  
\end{sloppypar}

This paper solidifies and extends CIU theory and relates it to currently popular methods in the domain called eXplainable AI (XAI). In recent years, the XAI domain has moved forward rapidly and has developed its own concepts and methods, which makes it difficult for current XAI researchers to understand and assess CIU in relation to their own work. The objectives of the paper are the following:
\begin{itemize}
    \item Present a solid mathematical theory for CIU.
    \item Provide distinct definitions of the concepts \textit{influence}, \textit{importance} and \textit{utility}.
    \item Define the new concept of \textit{contextual influence} derived from CIU.
    \item Situate CIU within the latest state-of-the-art in XAI and show that it performs better than core main-stream XAI methods. 
\end{itemize}

After this Introduction, Section \ref{sec:Background} goes through the theoretical constructs of CIU and relates CIU to the family of additive feature attribution methods such as Shapley values and LIME. Section \ref{sec:Experiments} provides empirical evidence for the theory in Section \ref{sec:Background}, followed by Conclusions.  

\section{Theory}\label{sec:Background}


\subsection{Additive Feature Attribution Methods}\label{Sec:ShapleyLIME_additive_fam}

We will here use notations from the paper by Lundberg and Lee \cite{NIPS2017_Lundberg_XAI} because it provides a unifying view on a whole family of outcome explanation methods called \textit{additive feature attribution (AFA)} methods. Such methods use an \textit{explanation model} $g$ that is an interpretable approximation of the original model $f$. 
The following definition is fundamental for AFA methods~\cite{NIPS2017_Lundberg_XAI}: 

\begin{definition} \textbf{AFA methods} have an explanation model that is a linear function of binary variables:
\begin{equation}
g(z')=\phi_{0}+\sum_{i=1}^{M}\phi_{i}z'_{i}, 
\label{Eq:AdditiveFeatureAttribution}
\end{equation}

where $z'\in \{0,1\}^M$, $M$ is the number of simplified input features, and $\phi \in \mathbb{R}$. 
\label{Def:AdditiveFeatureAttribution}
\end{definition}

Methods with explanation models $g$ matching this definition attribute an effect $\phi_{i}$ to each input feature. 
Since $\phi_{i}$ is a scalar, the definition signifies that the explanation model $g$ is linear by definition. 
Our interpretation of the variable $\phi_{i}$ is to call it `influence', which has also been used by other authors. Lundberg and Lee use the word `effect' for $\phi_{i}$ but they do also use the word `importance' with a similar meaning in \cite{NIPS2017_Lundberg_XAI}. 
It seems like most authors use `effect', `influence', `significance', `importance', etc. interchangeably. 

The \textit{Shapley value} is an AFA method originating from cooperative game theory \cite{shapley1953}. The concept was picked up by the XAI community \cite{StrumbeljKononenko_2010} and has become popular for producing outcome explanations following \cite{NIPS2017_Lundberg_XAI}, and the introduction of SHAP (SHapley Additive exPlanations). 
The method distributes the difference between the prediction output $f(x)$ and the \textit{reference level}\footnote{Called \textit{baseline} by many authors but `baseline' seems to be used also for other purposes. This is why we prefer using `reference level'.} $\phi_{0}$ to the input feature influences $\phi_{i}$ according to Equation \ref{Eq:AdditiveFeatureAttribution}. The most used $\phi_{0}$ value is the global average predicted value for the studied output in the training data set. If $f(x) > \phi_{0}$, then the sum of the terms $\phi_{i}$ must be positive, and vice versa if $f(x) < \phi_{0}$. 






\textit{Local Interpretable Model-agnostic Explanations (LIME)} is a popular AFA method, which creates a linear surrogate model $g$ that locally approximates the behaviour of the model to explain around the neighborhood of the instance being explained~\cite{ribeiro2016should}. 
The sign of $\phi_{i}$ determines if the influence of the input feature $i$ is negative or positive. The magnitude of $\phi_{i}$ expresses how great the influence is. 




\subsection{Decision Theory and Multi-Attribute Utility Theory}

In statistics, Decision Theory proposes a set of quantitative methods for reaching optimal, or at least rational, decisions. A decision problem must be capable of being formulated in terms of initial conditions and outcomes or courses of action, with their consequences. Each outcome is assigned a \textit{utility value} based on the \textit{preferences} of the decision maker(s). An optimal decision is one that maximizes the expected utility. 
It was proven already in 1947 that any individual whose preferences satisfy four axioms has a \textit{utility function}, $u$, by which an individual's preferences can be represented on an interval scale~\cite{vonneumann1947}.
If preferences over choices on attributes or \textit{input features} $1,\dots,n$ depend only on their marginal probability distributions, then the n-attribute utility function is additive according to:

\begin{equation}
u(x_{1},\dots ,x_{n})=\sum _{i=1}^{n}{k_{i}u_{i}(x_{i})}
\label{Eq:n_attribute_utility_function}
\end{equation}

where $u$ and the $u_{i}$ are normalized to the range [0,1], and the $k_{i}$ are normalization constants~\cite{Dyer2005}. If the goal is to simply rank-order the available choices, then a key condition for the additive form in Equation \ref{Eq:n_attribute_utility_function} is mutual \textit{preference independence}. 
A fundamental result in utility theory is that two attributes are additive-independent if and only if their two-attribute utility function is additive and has the form:

\begin{equation}
u(x_{1},x_{2})=u(x_{1}) + u(x_{2})
\label{Eq:additive_independent_utilities}
\end{equation}


CIU respects additive-independence as long as the underlying model is linear. However, the objective of CIU is not to provide a rank-ordering but providing an explanation for the outcome of an underlying DSS model, which requires and justifies breaking the additive-independence condition given in Equation \ref{Eq:additive_independent_utilities}, as explained in the next section. 

\subsection{Contextual Importance and Utility (CIU)}\label{Sec:CIU}


CIU estimates the values $k_{i}$ and $u_{i}(y_{i})$ in Equation~\ref{Eq:n_attribute_utility_function} for one or more input features $\{i\}$ in a specific context $C$ and any black-box model $f$, where the context is defined by the instance or situation to be explained. 

However, the use of Equation \ref{Eq:n_attribute_utility_function} makes it necessary to map output values $y=f(x)$ into utility values $u$ that are limited to the range $[0,1]$. In classification tasks, the $y$ values are usually probability values in the range $[0,1]$ by definition, so it can be considered that $u=y$. The same is not true for regression tasks. For instance, in the well-known Boston Housing data set, the output value is the median value of owner-occupied homes in \$1000's and is in the range $[5,50]$. A straightforward way of transforming that value into a utility value is an affine transformation $[5,50] \mapsto [0,1]$, assuming that the preference is to have a higher value. However, from a buyer's point of view, the preference might be for lower prices and then the transformation would rather be $[50,5] \mapsto [0,1]$. In this paper, we will assume that $u_{j}(y_{j})$ is an affine transformation of the form $u_{j}(y_{j})=Ay_{j}+b$, where $j$ is the output index. In practice, $u_{j}(y_{j})$ could have any shape as long as it produces values in the range $[0,1]$ but that case goes beyond the scope of the current paper (and theory). This takes us to the definition of \textit{Contextual Importance (CI)}.


\begin{definition}[Contextual Importance]
\begin{eqnarray}
CI_{j}(C,\{i\},\{I\})=\frac{umax_{j}(C,\{i\})-umin_{j}(C,\{i\})}{umax_{j}(C,\{I\})-umin_{j}(C,\{I\})},  
\label{Eq:GeneralCI}
\end{eqnarray}

where $\{i\} \subseteq \{I\}$ and $\{I\} \subseteq \{1,\dots,n\}$. $C$ is the instance/context to be explained and defines the values of input features that do not belong to $\{i\}$ or $\{I\}$. 
\end{definition}

For clarity, $\{i\}$ is the set of indices studied and $\{I\}$is the set of indices relative to which we calculate CI. When $\{I\}=\{1,\dots,n\}$, CI is calculated relative to the output utilities $u_{j}$. For instance, $CI_{j}(C,\{2\},\{1,\dots,n\})$ is the contextual importance of input $x_{2}$, whereas $CI_{j}(C,\{1,2,3\},\{1,\dots,n\})$ is the \textbf{joint} contextual importance of inputs $x_{1},x_{2},x_{3}$ and $CI_{j}(C,\{1,\dots,n\},\{1,\dots,n\})$ is the joint contextual importance of \textbf{all} inputs. $umin_{j}()$ and $umax_{j}()$ are the minimal and maximal utility values $u_{j}$ observed for output $j$ for all possible $x_{\{i\}}$ and $x_{\{I\}}$ values in the context $C$, while keeping other input values at $C$. Using $\{I\}\neq\{1,\dots,n\}$ makes it possible to also use and explain \textit{intermediate concepts} as described in \cite{FramlingAISB_1996,FramlingThesis_1996,Framling_XAI_WS_AAAI_2021}.

When $u_{j}(y_{j})=Ay_{j}+b$, then CI can be directly calculated as:

\begin{eqnarray}
CI_{j}(C,\{i\},\{I\})= \frac{ ymax_{j}(C,\{i\})-ymin_{j}(C,\{i\})}{ ymax_{j}(C,\{I\})-ymin_{j}(C,\{I\})}, 
\label{Eq:CI}
\end{eqnarray}

where $ymin_{j}()$ and $ymax_{j}()$ are the minimal and maximal $y_{j}$ values observed for output $j$. Equation~\ref{Eq:CI} is identical to the CI definitions in~\cite{FramlingThesis_1996,Framling_XAI_WS_AAAI_2021}.



The values of $umin_{j}$ and $umax_{j}$ can only be calculated exactly if the entire set of possible values for the input features $\{i\}$ is available and the corresponding $u_{j}$ values can be calculated in reasonable time. For categorical input features this is feasible as long as the number of possible values doesn't grow too big. For continuous-valued input features, using a \textit{Set of representative input vectors} $S(C,\{i\})$ is a model-agnostic approach to estimate $umin_{j}$ and $umax_{j}$. Algorithm \ref{algo:SRIV} shows how $S(C,\{i\})$ is constructed in the `ciu' R package \cite{Framling_XAI_WS_AAAI_2021}. The approach taken there is to limit the range of input values to intervals $[min_{\{i\}},max_{\{i\}}]$ for numerical input features. The studied instance $C$ is the first sample in $S(C,\{i\})$, followed by samples with the extreme values $min_{\{i\}}$ and $max_{\{i\}}$ for numerical input features $\{i\}$. For numerical input features, the remaining samples for achieving $N$ samples are generated randomly from the interval(s) $[min_{\{i\}},max_{\{i\}}]$. Other sampling methods could be envisaged, including model-specific ones like the one in Fr\"{a}mling's thesis \cite{FramlingThesis_1996} and remains a topic of future research. 

\begin{algorithm}[t]
\SetAlgoLined
\LinesNumbered
    \KwResult{$N \times M$ matrix $S(C,\{i\})$}
    \Begin{
    \ForAll{categorical input features}{
        $D \leftarrow$ all possible value combinations for discrete inputs $\{i\}$\;
        Randomize row order in $D$\;
        \If{$D$ has more rows than $N$}{Set $N$ to number of rows in $D$\;}
    }
    \ForAll{numerical input features}{
        Initialize $N \times M$ matrix $R$ with current input values $C$\;
        $R \leftarrow$ two rows per continuous-valued inputs in $\{i\}$ where the current value is replaced by the values $min_{\{i\}}$ and $max_{\{i\}}$ respectively\;
        $R \leftarrow$ fill remaining rows to $N$ with random values from intervals $[min_{\{i\}},max_{\{i\}}]$;
    }
    $S(C,\{i\}) \leftarrow$ concatenation of $C$ with merged $D$ and $R$, where $D$ is repeated if needed to obtain $N$ rows\;
    }
    \caption{Set of representative input vectors}
    \label{algo:SRIV}
\end{algorithm}

When the set of input features to explain $\{i\}$ is a subset of all input features $\{1,\dots,n\}$, then we apply the \textit{ceteris-paribus} principle, i.e. `other things held constant' for estimating their CI value. This signifies that all input features $\neg\{i\}$ are held constant at the values given by the studied instance $C$ while estimating $CI_{j}(C,\{i\})$ by varying the values of the input features $\{i\}$ according to Algorithm \ref{algo:SRIV}. This leads us to the following:

\begin{lemma}[Contextual Importance of input feature subsets $\{i\}$] \newline 
When $\{I\} \subseteq \{1,\dots,n\}$ and $\{i\} \subseteq \{I\} \Rightarrow$ \newline  $[umin_{j}(C,\{i\}),umax_{j}(C,\{i\})] \subseteq [umin_{j}(C,\{I\}),umax_{j}(C,\{I\})]$.
\label{The:SubsetTheorem}
\end{lemma}

\begin{proof}
When $\{i\} \subseteq \{I\}$, then $S(C,\{i\}) \subseteq S(C,\{I\})$ and \\$[umin_{j}(C,\{i\}),umax_{j}(C,\{i\})] \subseteq [umin_{j}(C,\{I\}),umax_{j}(C,\{I\})]$ \\when the number of samples $N \rightarrow \infty$.
\label{Proof:SubsetProof}
\end{proof}

When considering that $umax_{j}(C,\{i\}) - umin_{j}(C,\{i\}) \geq 0$, we get:

\begin{theorem}[Maximal range of Contextual Importance]
$CI_{j}(C,\{i\}) \in [0,1]$ for any set of input features $\{i\}$. 
\label{The:CI_Range}
\end{theorem}


The \textit{Contextual Utility (CU}) corresponds to the factor $u_{i}(x_{i})$ in Equation \ref{Eq:n_attribute_utility_function}. CU expresses to what extent the current value of a given input feature contributes to obtaining a high output utility $u_{j}$.  

\begin{definition}[Contextual Utility]
\begin{eqnarray}
CU_{j}(C,\{i\})=\frac{u_{j}(C)-umin_{j}(C,\{i\})}{umax_{j}(C,\{i\})-umin_{j}(C,\{i\})} 
\label{Eq:CU}
\end{eqnarray}
\end{definition}

When $u_{j}(y_{j})=Ay_{j}+b$, this can be written as:  

\begin{eqnarray}
CU_{j}(C,\{i\})=\left|\frac{ y_{j}(C)-yumin_{j}(C,\{i\})}{ymax_{j}(C,\{i\})-ymin_{j}(C,\{i\})}\right|, 
\label{Eq:CU_y}
\end{eqnarray}

where $yumin=ymin$ if $A$ is positive and $yumin=ymax$ if $A$ is negative. This definition of CU differs from CI definitions in \cite{FramlingThesis_1996,Framling_XAI_WS_AAAI_2021} by handling negative $A$ values correctly. 



\subsubsection{Illustration of Additive Independence in CIU for linear model $f$.}

We will next illustrate that CIU satisfies Equations \ref{Eq:n_attribute_utility_function} and \ref{Eq:additive_independent_utilities} when the input features $x_{1}, \dots, x_{n}$ are additive-independent. For this, we use the simple function $y=x_{1}+x_{2}$ with $x_{i}\in [0,1]$. In this case we can use $u_{i}(x_{i})=x_{i}$. Table \ref{Tab:WS_CIU} shows results for all the four zero-one combinations when 
$u(x_{i})=CI(x_{i})\times CU(x_{i})$. 

\begin{table}[ht]
\centering
\caption{Weighted sum input and output values, with corresponding CI and CU values.}
\begin{tabular}{|c|c|c|c|c|c|c|c|c|}
\hline
$x_{1}$ & $x_{2}$ & $y=x_{1} + x_{2}$& $CI(x_{1})$ & $CI(x_{2})$  & $CU(x_{1})$ & $CU(x_{2})$ & $u(x_{1})+u(x_{2})$ & $u(y)=u(x_{1},x_{2})$ \\
\hline
0 & 0 & 0 & 0.5 & 0.5 & 0 & 0 & 0 & 0 \\
0 & 1 & 1 & 0.5 & 0.5 & 0 & 1 & 0.5 & 0.5 \\
1 & 0 & 1 & 0.5 & 0.5 & 1 & 0 & 0.5 & 0.5 \\
1 & 1 & 2 & 0.5 & 0.5 & 1 & 1 & 1 & 1 \\
\hline
\end{tabular}
\label{Tab:WS_CIU}
\end{table}

We now go to the core point of disruption of CIU with utility theory: most models $f$ for which we would like to provide explainability are non-linear and their input features tend to be dependent on each other. Therefore, CIU proposes to abandon the requirement of additive-independence of Equation~\ref{Eq:additive_independent_utilities}. An initial assumption of CIU is indeed that both the importance and the utility function can (and usually do) depend on the values of other input features, which is the main reason for using the word \textit{contextual} in CIU. 


In order to illustrate the need and necessity to take the contextual aspects into account for outcome explanation, we will study how to explain results of simple OR and XOR functions, where the input features are clearly dependent. The results are shown in Tables \ref{Tab:OR_CIU} and \ref{Tab:XOR_CIU}.

\begin{table}[ht]
\centering
\caption{OR function input and output values, with corresponding CI and CU values.}
\begin{tabular}{|c|c|c|c|c|c|c|c|c|}
\hline
$x_{1}$ & $x_{2}$ & $y=x_{1} \vee x_{2}$& $CI(x_{1})$ & $CI(x_{2})$  & $CU(x_{1})$ & $CU(x_{2})$ & $u(x_{1})+u(x_{2})$ & $u(y)=u(x_{1},x_{2})$ \\
\hline
0 & 0 & 0 & 1 & 1 & 0 & 0 & 0 & 0 \\
0 & 1 & 1 & 0 & 1 & NaN & 1 & 1 & 1 \\
1 & 0 & 1 & 1 & 0 & 1 & NaN & 1 & 1 \\
1 & 1 & 1 & 0 & 0 & NaN & NaN & 0 & 1 \\
\hline
\end{tabular}
\label{Tab:OR_CIU}
\end{table}

\begin{table}[ht]
\centering
\caption{XOR function input and output values, with corresponding CI and CU values.}
\begin{tabular}{|c|c|c|c|c|c|c|c|c|}
\hline
$x_{1}$ & $x_{2}$ & $y=x_{1} \oplus x_{2}$& $CI(x_{1})$ & $CI(x_{2})$  & $CU(x_{1})$ & $CU(x_{2})$ & $u(x_{1})+u(x_{2})$ & $u(y)=u(x_{1},x_{2})$ \\
\hline
0 & 0 & 0 & 1 & 1 & 0 & 0 & 0 & 0 \\
0 & 1 & 1 & 1 & 1 & 1 & 1 & 2 & 1 \\
1 & 0 & 1 & 1 & 1 & 1 & 1 & 2 & 1 \\
1 & 1 & 0 & 1 & 1 & 0 & 0 & 0 & 0 \\
\hline
\end{tabular}
\label{Tab:XOR_CIU}
\end{table}

A core reason for showing these three simple examples is to emphasize that both CI and CU are \textbf{absolute} values in the range $[0,1]$, as opposed to \textbf{relative} values used by AFA methods. $CI_{j}(C,\{i\})=0$ signifies that in the context $C$ the input feature(s) $\{i\}$ have no effect on the utility $u_{j}$ of output $j$. $CI_{j}(C,\{i\})=1$ signifies that changes to the values of input feature(s) $\{i\}$ can modify the value of $u_{j}$ over the entire range $[0,1]$. Similarly, $CU_{j}(C,\{i\})=0$ signifies that the value(s) of input feature(s) $\{i\}$ are the least favorable (in the sense of utility $u_{j}$) for the output $j$. $CU_{j}(C,\{i\})=1$ signifies that the value(s) of input feature(s) $\{i\}$ are the most favorable for the output $j$.

\subsection{Contextual influence}

CI and CU produce explanations from any model $f$ in a uniform way, no matter if the model is linear or not, continuous-valued or discrete, handcoded or created via machine learning. However, in order to compare CIU with AFA methods, we define \textit{Contextual influence}. 
We begin by a contextual version of the term $k_{i}u_{i}(x_{i})$ in Equation \ref{Eq:n_attribute_utility_function}: 

\begin{equation*}
    Cinfluence_{j}(C,\{i\}) = CI_{j}(C,\{i\})\times CU_{j}(C,\{i\}),
\label{Eq:Cinfluence}
\end{equation*}

when $k_{i}=CI_{j}(C,\{i\})$ and $u_{i}=CU_{j}(C,\{i\})$. 
$Cinfluence$ can be scaled into any range $[rmin,rmax]$, which leads us to the following definition:

\begin{definition}[Contextual influence]
\begin{equation}
\phi=(rmax-rmin) \times CI \times (CU - neutral.CU)
\label{Eq:scaled_influence}
\end{equation}

where `$_{j}(C,\{i\})$' has been omitted from all three terms $\phi$, $CI$, and $CU$ for easier readability. 
\end{definition}

The symbol $\phi$ has been chosen on purpose to signify `influence' as for Shapley values and LIME. We use $[rmin,rmax]=[-1,1]$ in Section \ref{sec:Experiments}. Setting $neutral.CU=0.5$ restricts $\phi$ values to only negative, zero or positive, as for Shapley values and LIME. 








\subsection{CIU versus Additive Feature Attribution Methods}

As shown in the previous Sections, CIU makes a clear distinction between `importance' and `influence'. Furthermore, CIU uses the notions of `utility function' and `utility', which are not considered by any known AFA method. As shown by the following differences, \textbf{CIU is not an AFA method}:

\begin{itemize}
    \item CIU does not use or create any explanation model $g$.
    \item CI and CU can be used for calculating an influence measure $\phi$ but it is not possible to do it the other way around. 
    \item CI and CU provide absolute values in the range $[0,1]$ that have precise definitions, whereas $\phi$ values express relative influence between input features.
    \item CIU is defined using utility theory and CIU  explanations are entirely based on elements of that theory. CIU does not attempt to mimic or approximate the original function $f$ in any way. 
    \item CIU has no notion $z'_{i}$ of presence or not of an input feature and should not be confused with so-called occlusion-based methods~\cite{ZeilerFergus_2014}. 
\end{itemize}

Intuitively, it might be possible to consider Contextual influence in Equation \ref{Eq:scaled_influence} to be an AFA method, which is one reason for using the symbol $\phi$ for it. However, the fact that the reference level $neutral.CU$ is defined on utility values $u_{j}$ and not on output values $y_{j}$ as in AFA methods is a major difference. Furthermore, there's no additivity requirement on Contextual influence, even though additivity could be imposed by normalization. Still, further research on comparing Contextual influence and AFA methods is interesting and ongoing. 





\section{Experimental Evaluation}\label{sec:Experiments}

\begin{figure}
\centering
\begin{subfigure}{.3\columnwidth}
\centering
\includegraphics[width=\textwidth]{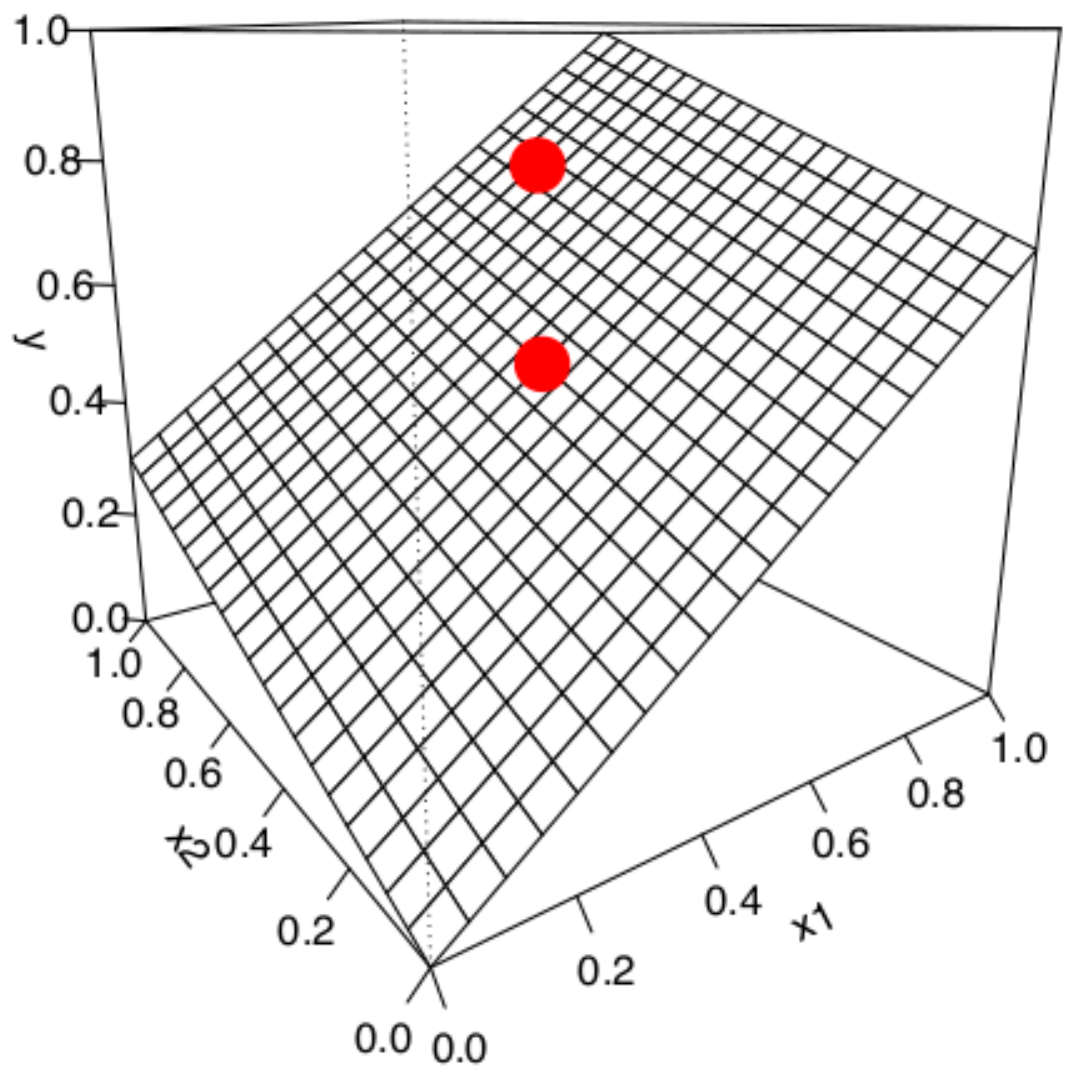}
\caption{Weighted sum.} \label{Fig:linear_func}
\end{subfigure}
\begin{subfigure}{.3\columnwidth}
\centering
\includegraphics[width=\textwidth]{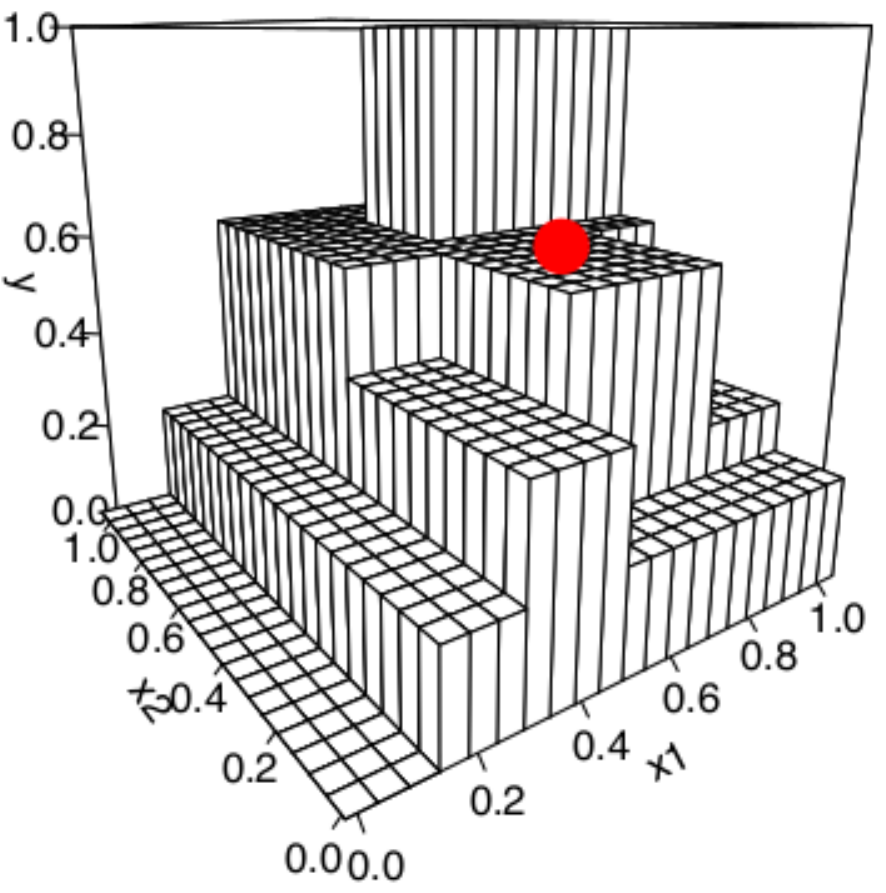}
\caption{Rule-based.} \label{Fig:rule_func}
\end{subfigure}
\begin{subfigure}{.3\columnwidth}
\centering
\includegraphics[width=\textwidth]{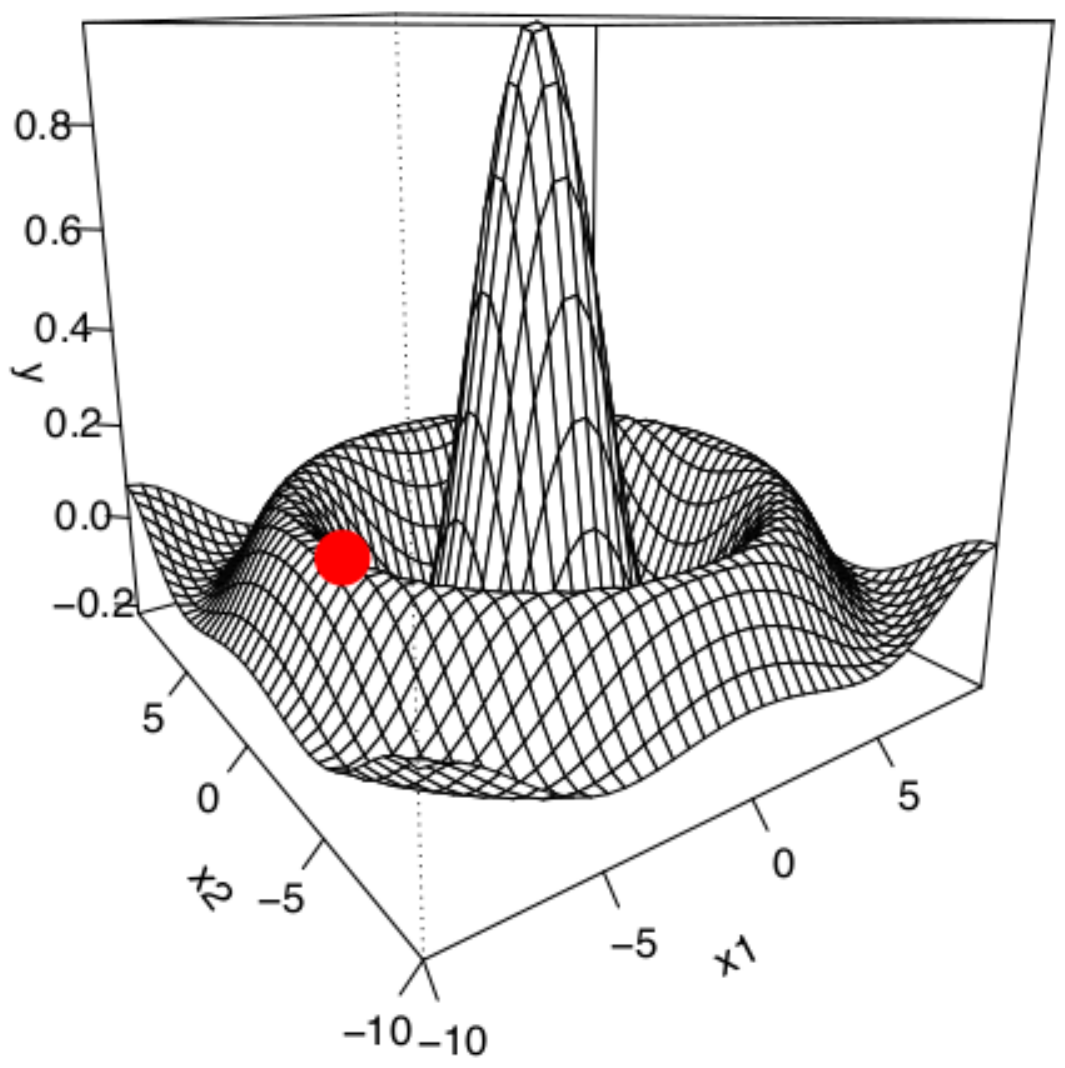}
\caption{Sombrero.} \label{Fig:sombrero}
\end{subfigure}
\caption{Linear, rule-based and non-linear models used in the study. 
}
\label{Fig:TestFunctions}
\end{figure}
\begin{figure}[t]
\centering
\includegraphics[width=0.32\textwidth]{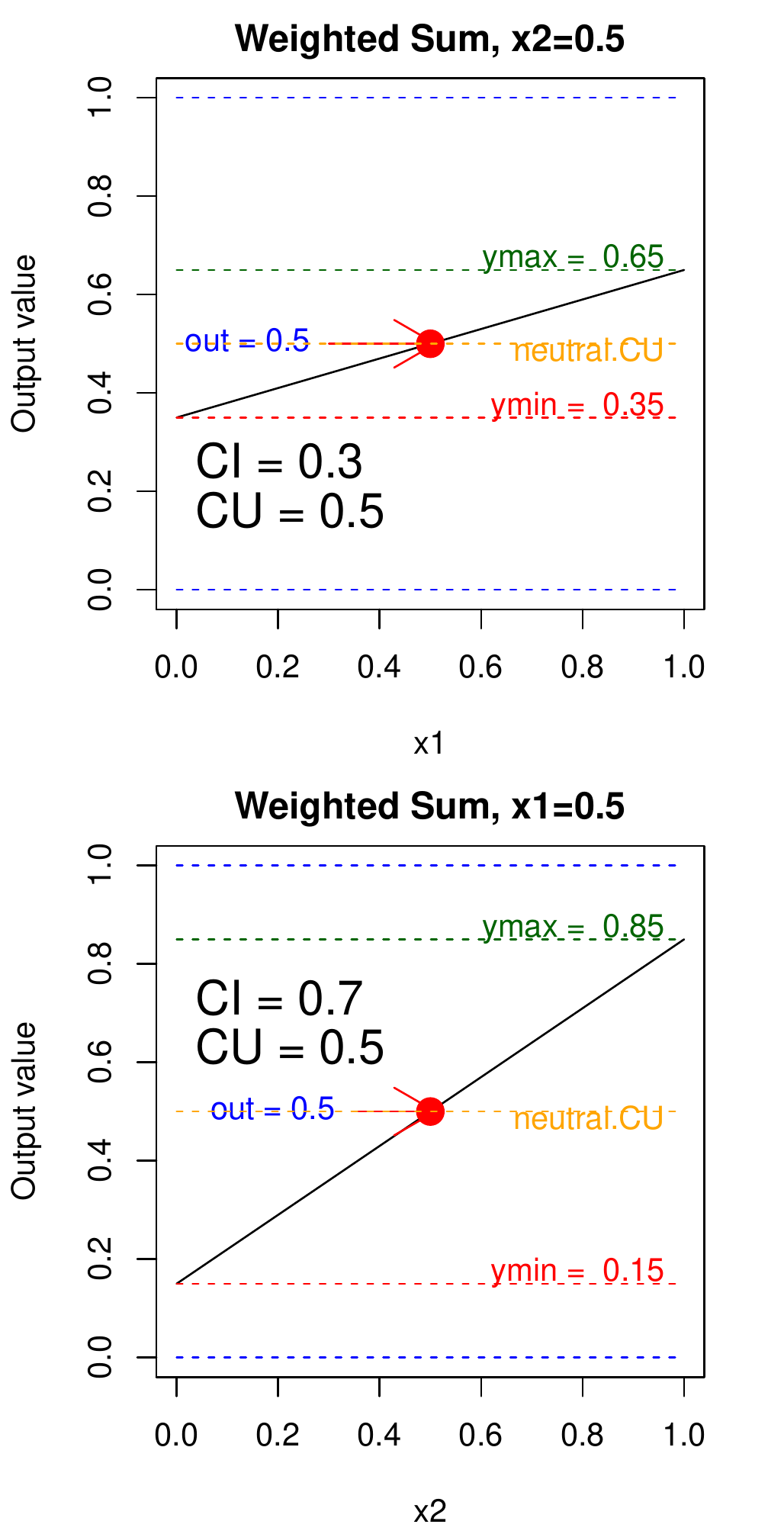}
\includegraphics[width=0.32\textwidth]{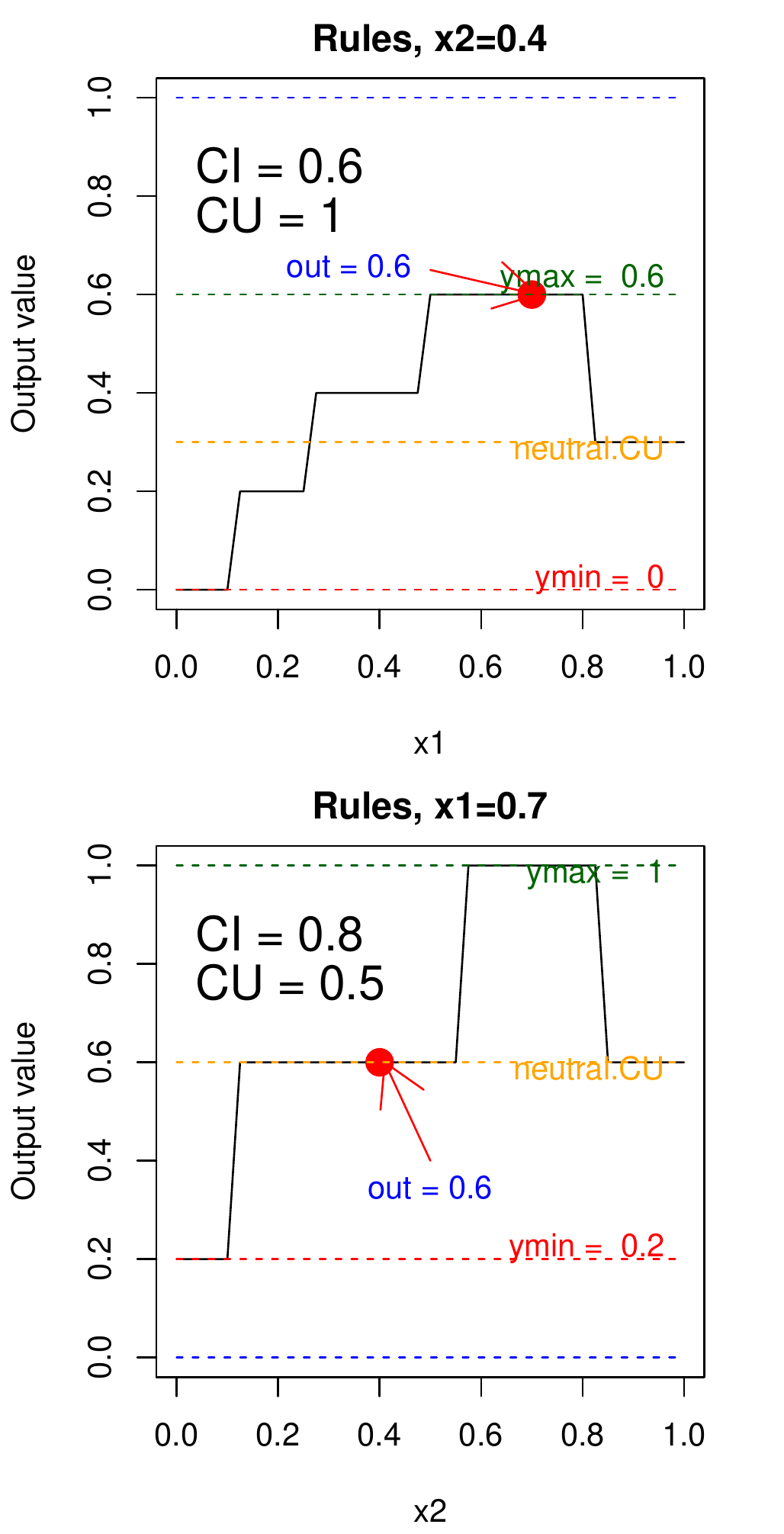}
\includegraphics[width=0.32\textwidth]{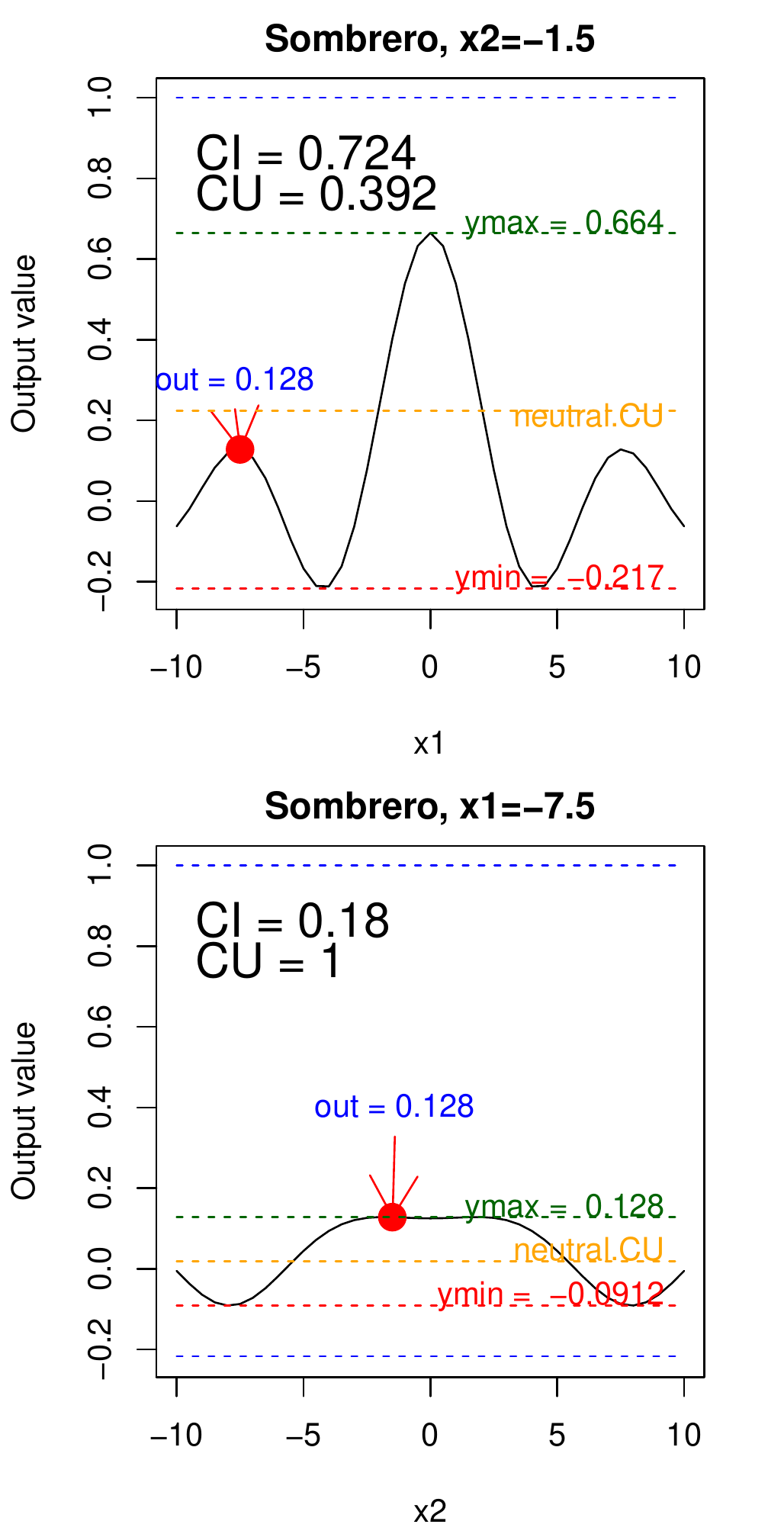}
\caption{Output value as a function of one variable for the test functions, with illustration of how CI and CU are calculated.} \label{Fig:X1X2_CIU}
\end{figure}

In this section we compare CIU, contextual influence, Shapley values and LIME for three known functions that have two input features $x_{1}, x_{2}$ and one output value $y$. The functions are linear ($y = 0.3x_1 + 0.7x_2$), rule-based and `sombrero' ($y = \sin(\sqrt{x_1^2 + x_2^2})/\sqrt{x_1^2 + x_2^2}$), as shown in Figure \ref{Fig:TestFunctions}. The studied input values $C=(x_{1},x_{2})$ are indicated by the red dots in Figure \ref{Fig:TestFunctions}. Figure \ref{Fig:X1X2_CIU} shows how the output $y$ changes as a function of one input feature while keeping constant the value of the other input feature, together with values and illustrations of CIU calculations. 

\begin{figure}[h]
\centering
\begin{tabular}{cccc}
\includegraphics[width=0.24\textwidth]{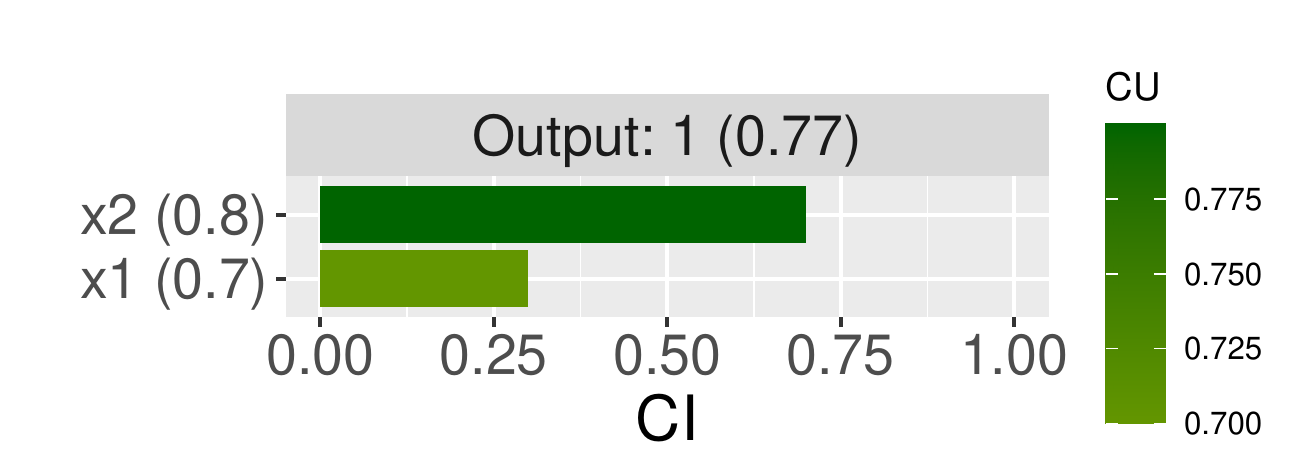} &
\includegraphics[width=0.24\textwidth]{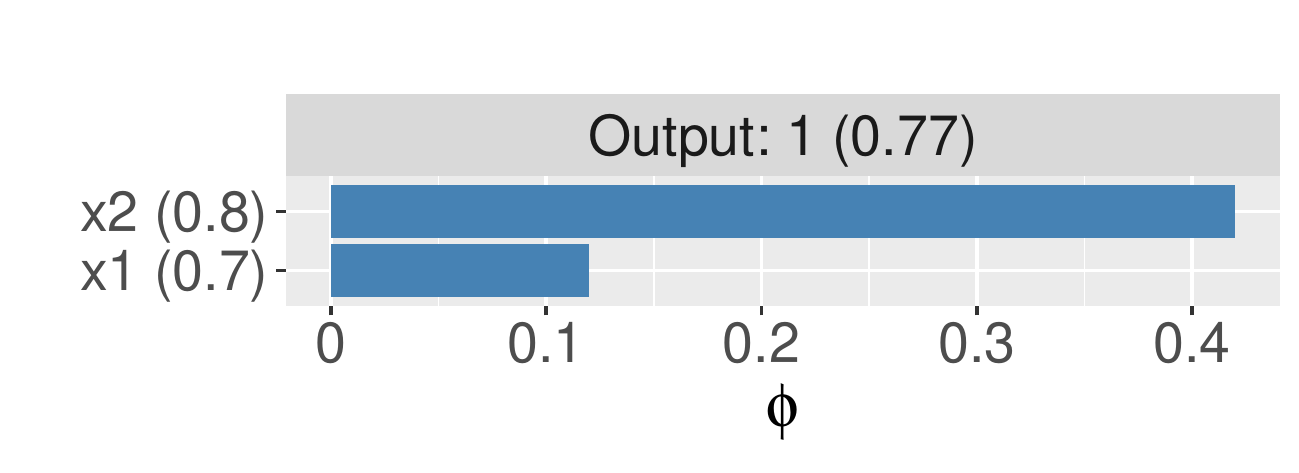} &
\includegraphics[width=0.24\textwidth]{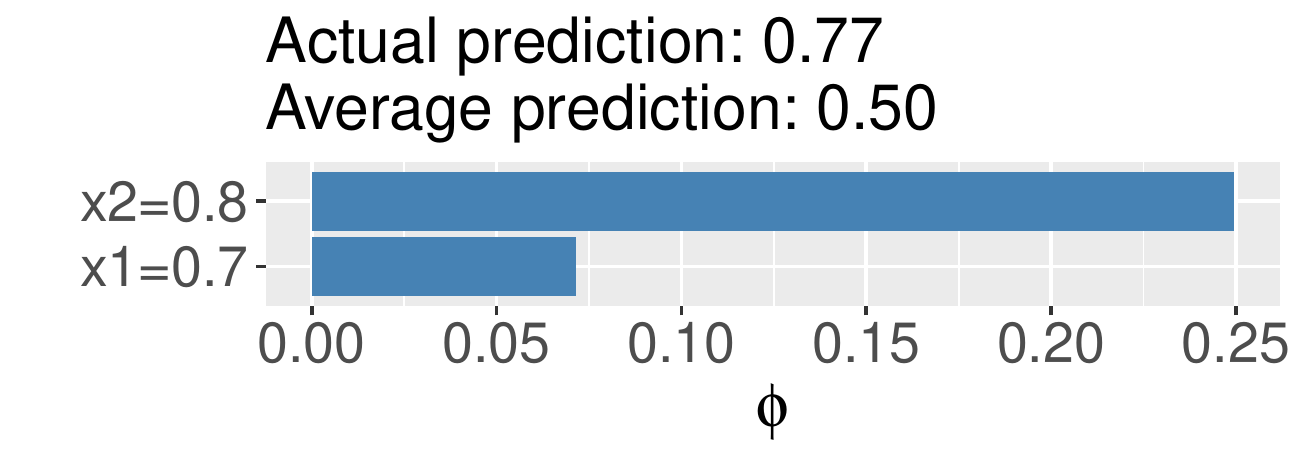} &
\includegraphics[width=0.24\textwidth]{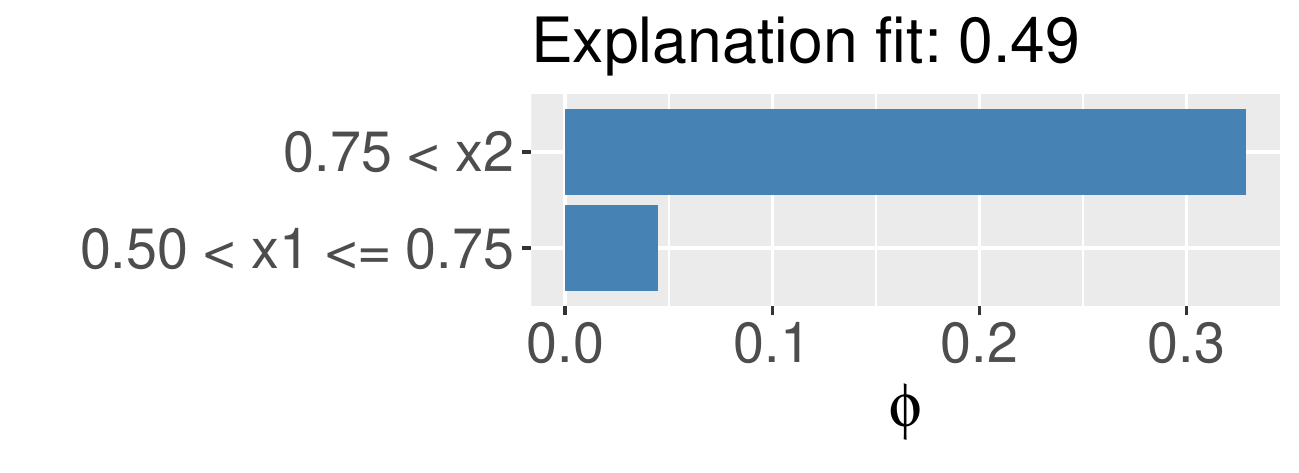} \\
\includegraphics[width=0.24\textwidth]{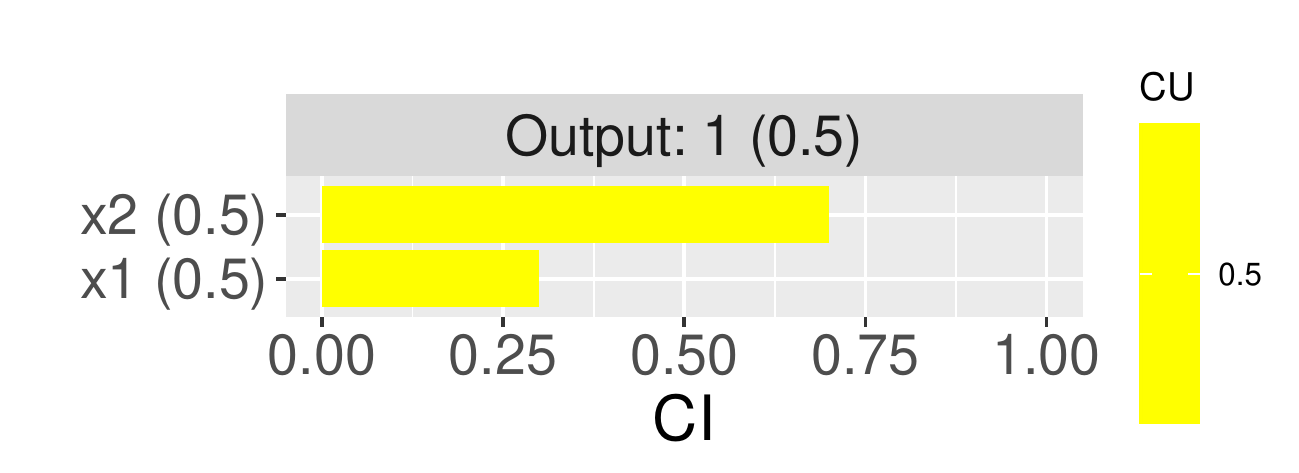} &
\includegraphics[width=0.24\textwidth]{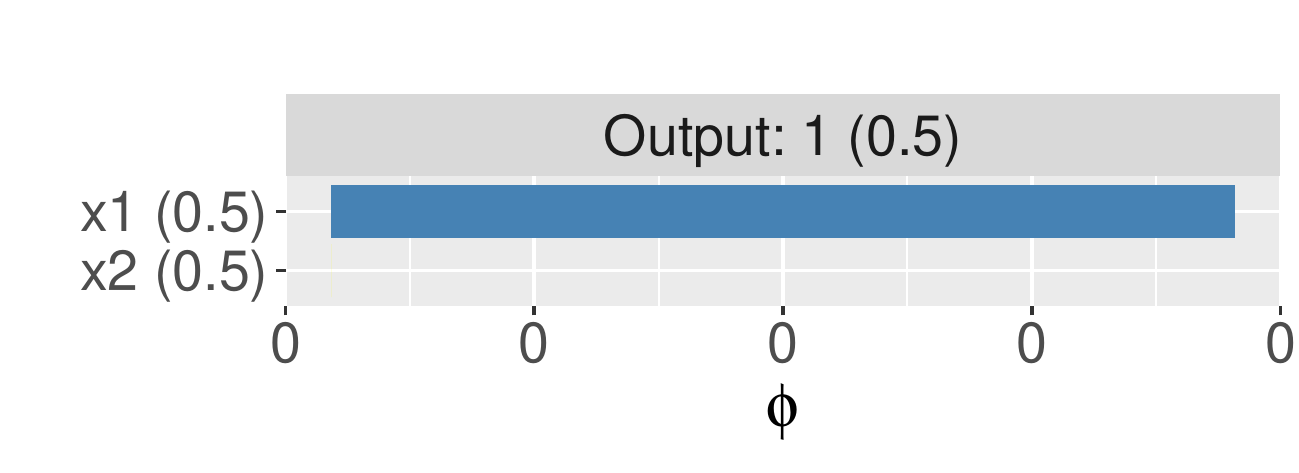} &
\includegraphics[width=0.24\textwidth]{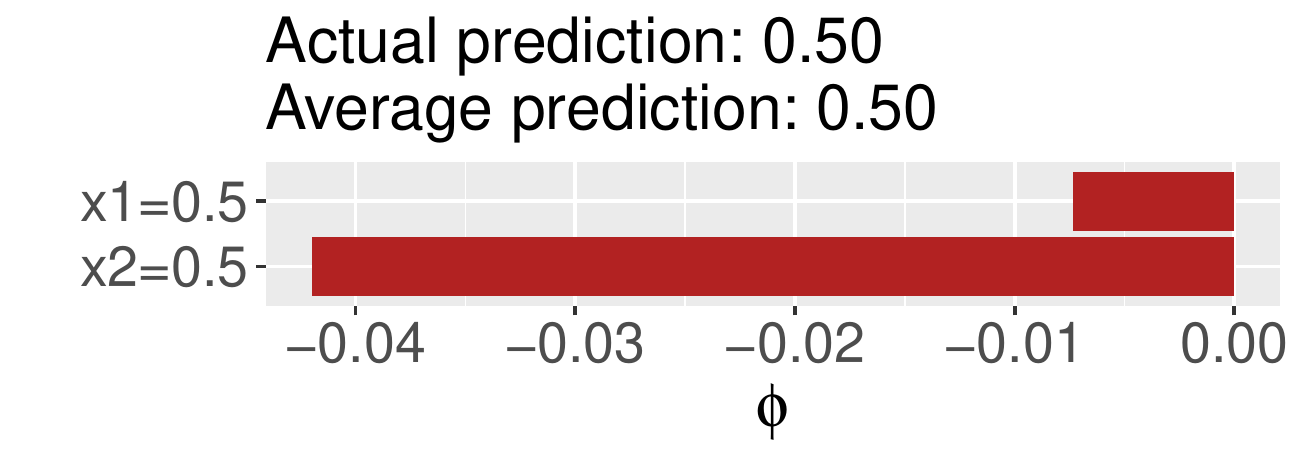} &
\includegraphics[width=0.24\textwidth]{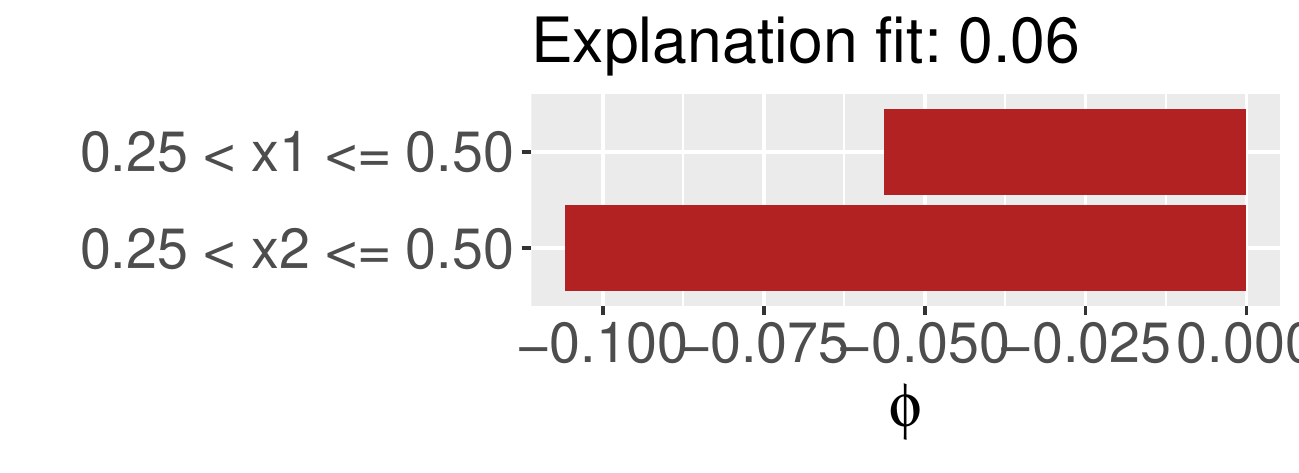} \\
\includegraphics[width=0.24\textwidth]{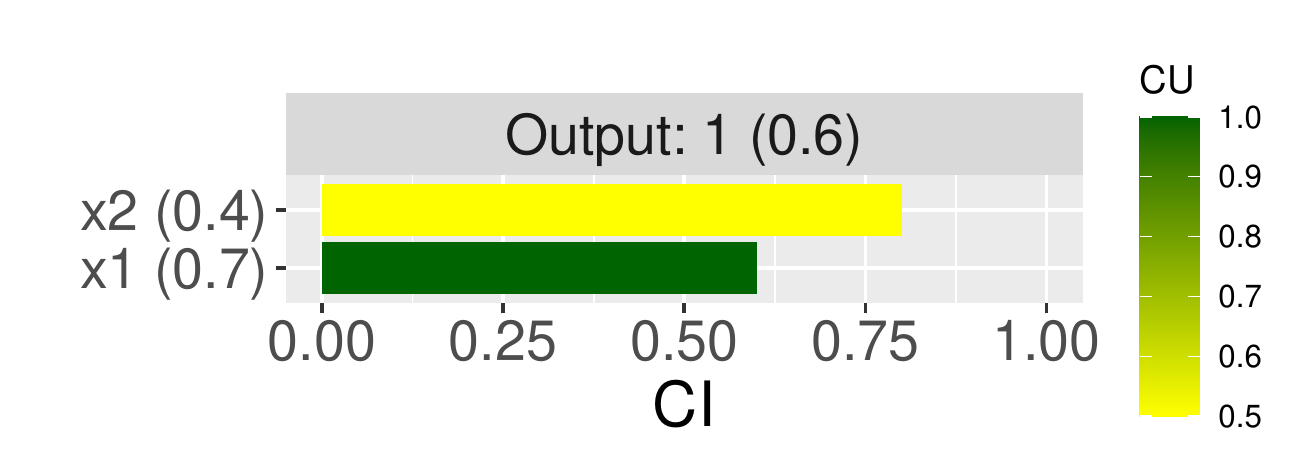} &
\includegraphics[width=0.24\textwidth]{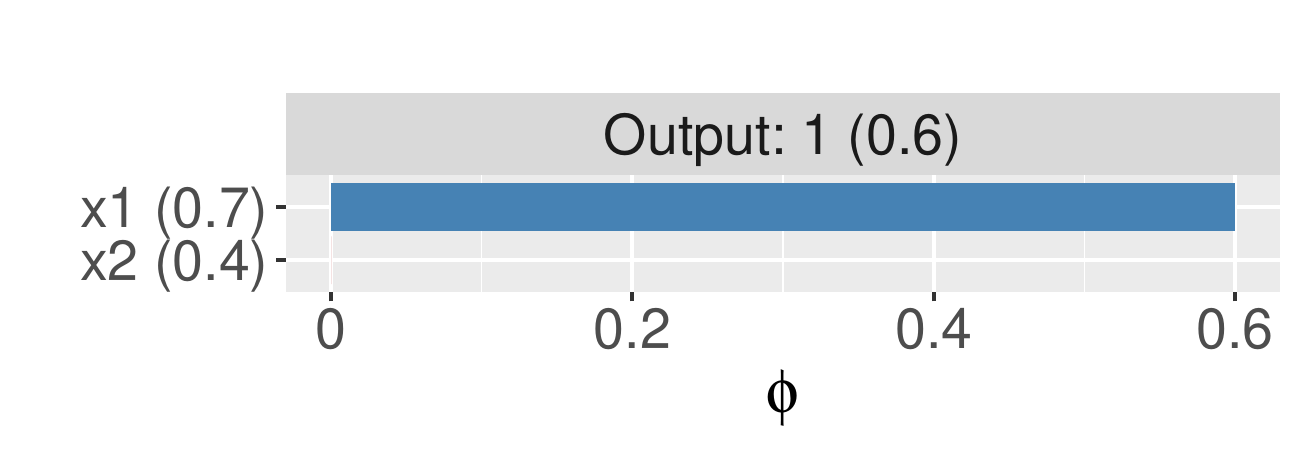} &
\includegraphics[width=0.24\textwidth]{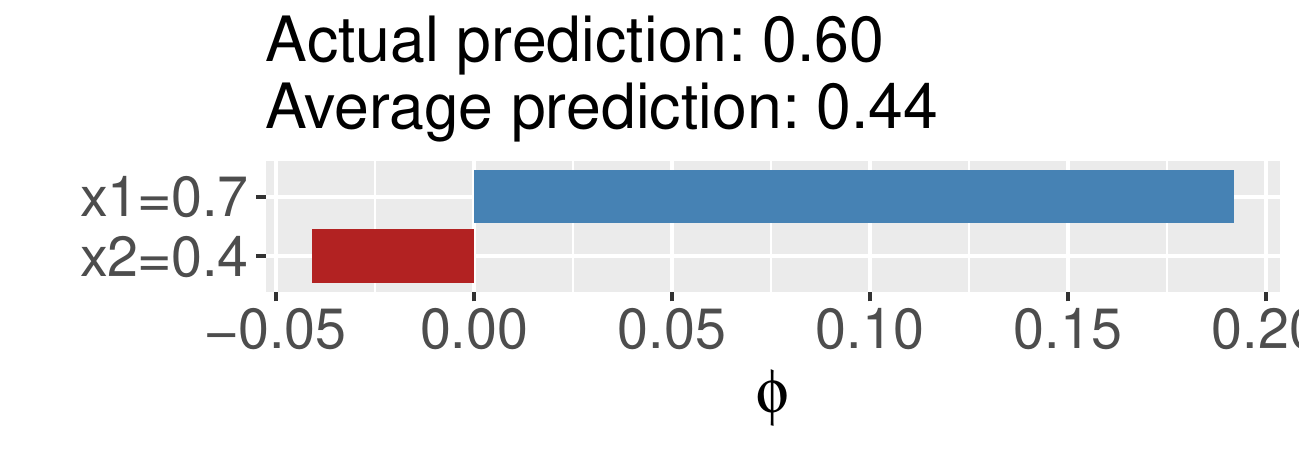} &
\includegraphics[width=0.24\textwidth]{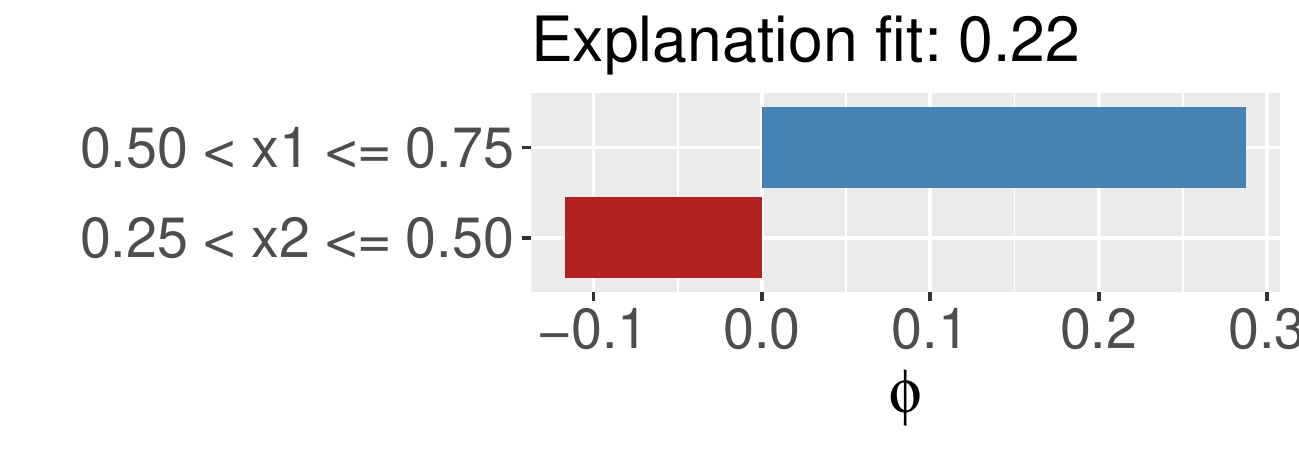} \\
\includegraphics[width=0.24\textwidth]{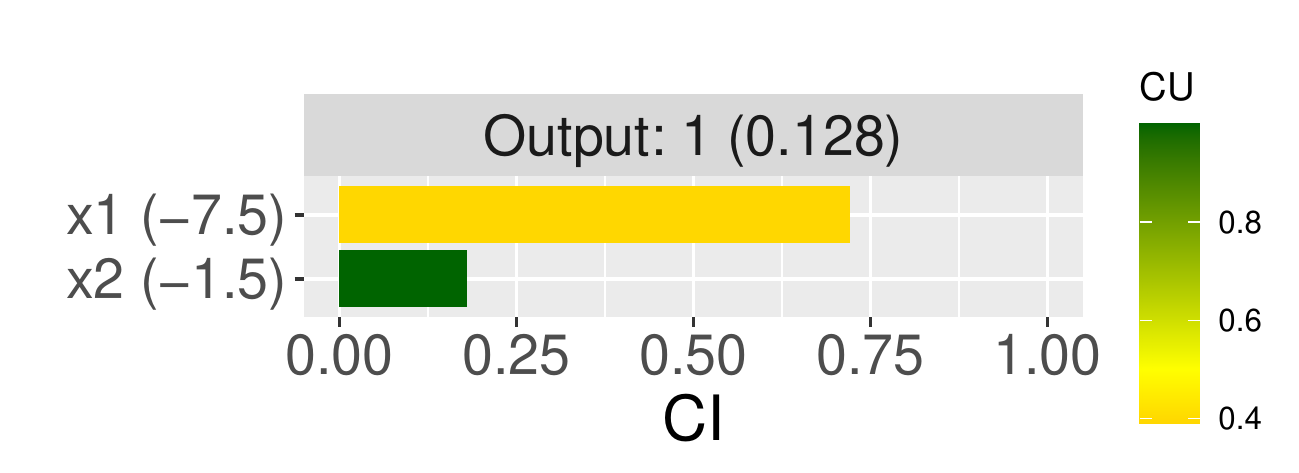} &
\includegraphics[width=0.24\textwidth]{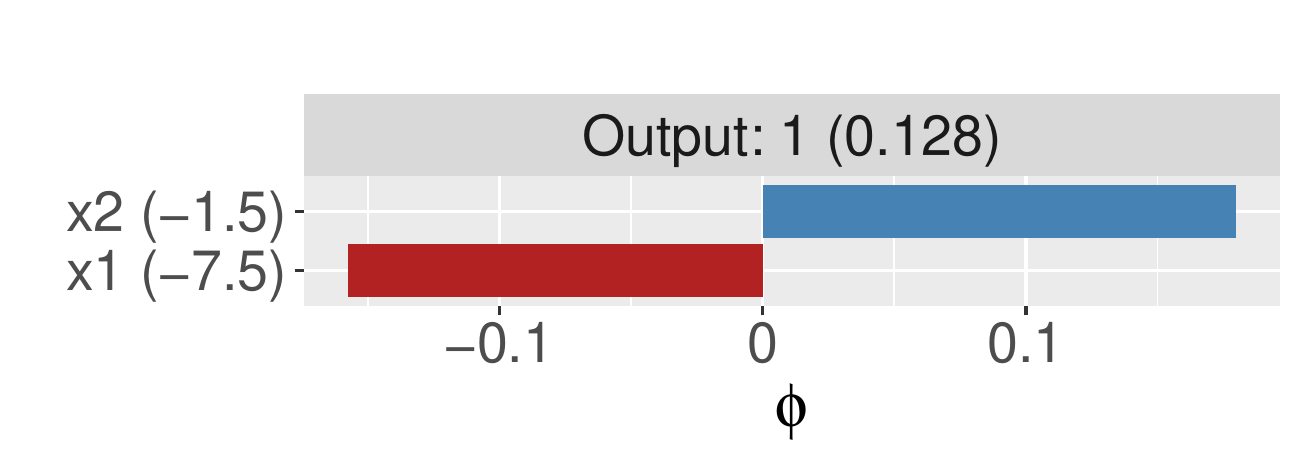} &
\includegraphics[width=0.24\textwidth]{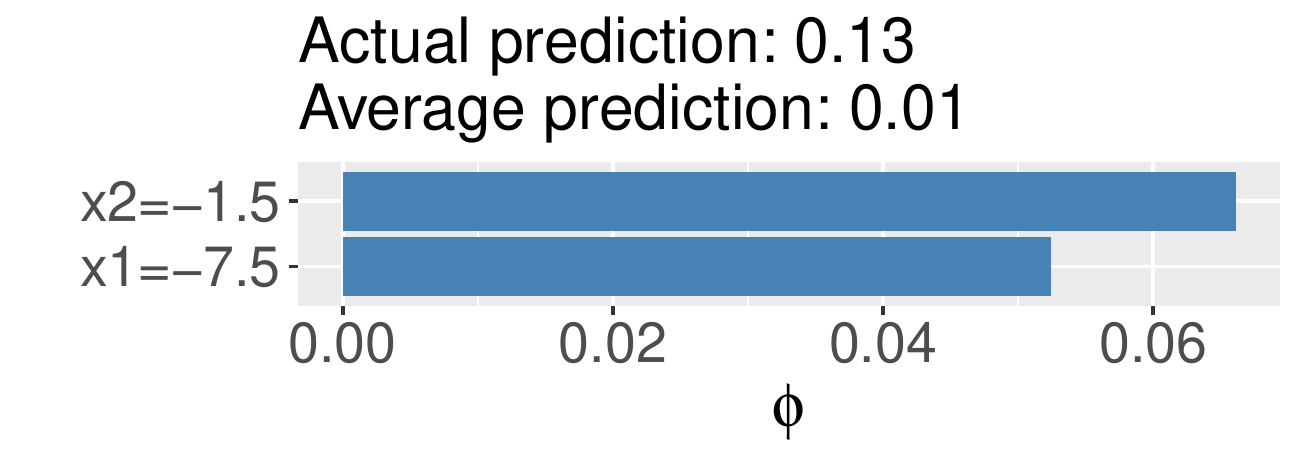} &
\includegraphics[width=0.24\textwidth]{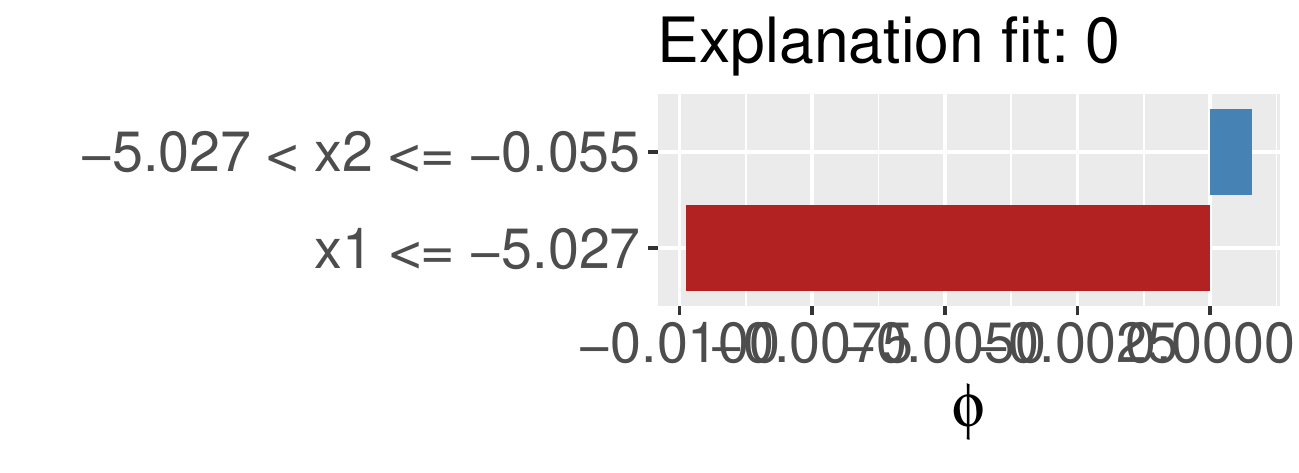} \\
CIU & Contextual influence & Shapley & LIME \\
\end{tabular}
\caption{Bar plots for the four methods on the four test functions.}
\label{Fig:XAI_barplots}
\end{figure}

CIU results are produced using the `ciu' R package \cite{Framling_XAI_WS_AAAI_2021}. Shapley values are produced with the `IML' R package \cite{MolnarIML_2018} and 
LIME results are produced with the `lime' R package~\cite{PedersenLimeR_2019}. All methods were run with default parameters ($N=100$ for CIU). In Figure \ref{Fig:XAI_barplots} the order of input features is determined automatically by the respective package, so it is not necessarily the same in all bar plots.  

\begin{table}[b]
\centering
\caption{Results for known functions with two inputs and one output.}
\begin{tabular}{|c|c|c|c|c|c|c|c|c|c|c|c|c|c|}
\hline
$f(x)$ & $x_{1}$ & $x_{2}$ & $y$ & $CI_{1}$ & $CI_{2}$  & $CU_{1}$ & $CU_{2}$ & $\phi_{1}^{ciu} $ & $\phi_{2}^{ciu}$ & $\phi_{1}^{shap} $ & $\phi_{2}^{shap}$ & $\phi_{1}^{lime} $ & $\phi_{2}^{lime}$ \\
\hline
Linear & 0.7 & 0.8 & 0.77 & 0.3 & 0.7 & 0.7 & 0.8 & 0.12 & 0.42 & 0.065 & 0.208 & 0.040 & 0.331 \\
Linear & 0.5 & 0.5 & 0.5 & 0.3 & 0.7 & 0.5 & 0.5 & 0.0 & 0.0 & 0.007 & -0.021 & -0.054 & -0.097 \\
Rules & 0.7 & 0.4 & 0.6 & 0.6 & 0.8 & 1.0 & 0.5 & 0.6 & 0.0 & 0.218 & -0.046 & 0.285 & -0.117  \\
Sombrero & -7.5 & -1.5  & 0.128 & 0.724 & 0.18 & 0.392 & 0.998 & -0.157 & 0.18 & 0.061 & 0.032 & -0.019 & 0.010 \\
\hline
\end{tabular}
\label{Tab:SimpleF_results}
\end{table}

CIU can use the test functions directly as the model to explain and does not need a data set. Both Shapley values and LIME do require a training data set, which has been generated as a regular grid of points $(x_{1},x_{2})$. A grid $x_{1},x_{2} \in [0,1]$ with step size $0.05$ is used for the linear and rule-based functions. For the `sombrero' function, $x_{1},x_{2} \in [-10,10]$ with step size $0.51$. Table \ref{Tab:SimpleF_results} and Figure \ref{Fig:XAI_barplots} show the results of the different methods. The following observations are made:

\textit{CIU} consistently describes a) how much the utility (`goodness') of $y$ can change when modifying the value of $x_{i}$ from the least favorable to the most favorable, and b) how favorable the current value $x_{i}$ is for the utility  $u(y)$, for the current instance/context $C$. CIU shows complete fidelity towards the underlying model, as indicated in Table \ref{Tab:SimpleF_results}, which correspond exactly to the weights and input/utility values for the linear function. CI and CU values can also be `seen' and understood directly from the graphs in Figure \ref{Fig:X1X2_CIU}.

\textit{Influence-based} explanations are inconsistent between the different methods for the three last test cases. In particular for the linear function with $(x_{1},x_{2})=(0.5,0.5)$, the influence-based explanations are in-existent because all $\phi_{i}$ values are (or should be) zero, which is indeed the case for all three influence-based methods. The slight deviations from zero are only due to numerical imprecision for contextual influence, whereas sampling leads to stochastic $\phi$ values that are normally distributed  around zero for Shapley values and LIME. 


\textit{Contextual influence} shows greater stability than the other influence-based methods. It also remains possible to `read' and understand the contextual influence values directly from Figure \ref{Fig:X1X2_CIU} because it corresponds to $y - neutral.CU$ relative to the $[ymin_{i},ymax_{i}]$ range. 

\textit{Shapley values} distribute the difference $f(x)-\phi_{0}$ `fairly' over all inputs $\phi_{i}$. However, when that difference is zero, as for the linear function with $(x_{1},x_{2})=(0.5,0.5)$, then there is nothing to distribute, at least not to input features with identical and `average' values, so Shapley values fail in producing any explanation in this case. The corresponding bar plot in Figure \ref{Fig:XAI_barplots} gives an impression that there is a successful explanation but that is because influence values are relative, so the scale of the $x$-axis is extended. In practice, both $\phi$ values are (or should be) zero, as seen in Table \ref{Tab:SimpleF_results}. 

\textit{LIME}. For some reason, LIME's `Explanation fit' is 0.5 for the linear function with $(x_{1},x_{2})=(0.7,0.8)$ but only 0.06 with $(x_{1},x_{2})=(0.5,0.5)$. It also remains unclear what is actually the reference level $\phi_{0}$ used by LIME. 

The experiments shown in this paper emphasize the theoretical difference between CIU and AFA methods, as well as the difference between the concepts of importance, utility and influence. All methods are applicable to any real-world tabular data sets and an extensive comparison between CIU, LIME and Shapley values is presented in \cite{FramlingEtAL_CIU_LIME_SHAP_EXTRAAMAS_2021}. The CIU Github site \url{https://github.com/KaryFramling/ciu} provides executable examples at least for the well-known benchmark data sets Iris, Boston, Heart Disease, UCI Cars, Diamonds, Titanic and Adult and several different machine learning models. CIU is also implemented for image explanations as reported in \cite{Framling_CIU_image_EXTRAAMAS_2021,KnapicMalhiSalujaFramling_MAKE_gastroimages_2021}. The source code used in this paper is published at \url{https://github.com/KaryFramling/AJCAI_2021}.

\section{Conclusion}

This paper extends the theory of CIU beyond the original theory in \cite{FramlingThesis_1996} and defines the new concept of contextual influence. CIU is compared to current state-of-the-art methods, notably the family of AFA methods. As shown in the paper, CIU provides new flexibility and expressiveness by separating the notions of \textit{importance} and \textit{utility} from the notion of \textit{influence} used by AFA methods. It is also illustrated why `influence' alone lacks in explanation capability.
Identified advantages of CIU compared to AFA methods are:

\begin{enumerate}
    \item CI and CU provide absolute values in the $[0,1]$ range that have clear definitions and interpretations
    \item Separate `importance' and `utility' concepts allow for more fine-grained and accurate explanations than `influence' alone.
    \item CIU has only one `tunable' parameter (the number $N$ of samples to use), which provides robustness and simplicity of use. 
    \item CIU does not need access to a data set.
    \item CIU is not a `black box' in itself because CI and CU values can be `read out' directly from input versus output graphs.  
    \item CIU's faithfulness/fidelity towards the model $f$ is guaranteed because no interpretable model $g$ is needed. CIU's faithfulness only depends on how accurately $umin$ and $umax$ can be estimated. 
    \item CIU has a solid and proven mathematical background and framework in multi-attribute utility theory, which puts it at least on the same level of rigor as Shapley values. 
\end{enumerate}

\bibliographystyle{splncs04}
\bibliography{AJCAI_2021}

\end{document}